%% file: KDD_paper.tex
\documentclass[sigconf]{acmart}

\usepackage{subfigure}
\usepackage{diagbox}

\usepackage{tikz}
\usetikzlibrary{fit}	
\usetikzlibrary{backgrounds}	% drawing the background after the foreground

\usepackage{multirow} 
\usepackage{booktabs}

\usetikzlibrary{%
  arrows,%
  shapes.misc,% wg. rounded rectangle
  shapes.arrows,%
  chains,%
  matrix,%
  positioning,% wg. " of "
  scopes,%
  decorations.pathmorphing,% /pgf/decoration/random steps | erste Graphik
  shadows%
}
\usepackage{pgfplots}
\pgfplotsset{compat=1.14}

\AtBeginDocument{%
  \providecommand\BibTeX{{%
    \normalfont B\kern-0.5em{\scshape i\kern-0.25em b}\kern-0.8em\TeX}}}

\begin{document}

\title{A Generic Object Re-identification System for Short Videos}

\author{Tairu Qiu}
\email{18210240156@fudan.edu.cn}
\affiliation{%
  \institution{Fudan University}
%   \streetaddress{Street Address}
%   \city{City}
  \state{Shanghai}
  \country{China}
}

\author{Guanxian Chen}
\email{18210240053@fudan.edu.cn}
\affiliation{%
  \institution{Fudan University}
%   \streetaddress{Street Address}
%   \city{City}
  \state{Shanghai}
  \country{China}
}

\author{Zhongang Qi}
\email{zhongangqi@tencent.com}
\affiliation{%
  \institution{Applied Research Center (ARC)}
%   \streetaddress{Street Address}
  \city{PCG}
  \state{Tencent}
  \country{China}
}

\author{Bin Li}
\authornotemark[1]
\email{libin@fudan.edu.cn}
\affiliation{%
  \institution{Fudan University}
%   \streetaddress{Street Address}
%   \city{City}
  \state{Shanghai}
  \country{China}
}

\author{Ying Shan}
\email{yingsshan@tencent.com}
\affiliation{%
  \institution{Applied Research Center (ARC)}
%   \streetaddress{Street Address}
  \city{PCG}
  \state{Tencent}
  \country{China}
}

\author{Xiangyang Xue}
\email{xyxue@fudan.edu.cn}
\affiliation{%
  \institution{Fudan University}
%   \streetaddress{Street Address}
%   \city{City}
  \state{Shanghai}
  \country{China}
}


\begin{abstract}
Short video applications like TikTok and Kwai have been a great hit recently. In order to meet the increasing demands and take full advantage of visual information in short videos, objects in each short video need to be located and analyzed as an upstream task. A question is thus raised -- how to improve the accuracy and robustness of object detection, tracking, and re-identification across tons of short videos with hundreds of categories and complicated visual effects (VFX). 
To this end, a system composed of a detection module, a tracking module and a generic object re-identification module, is proposed in this paper, which captures features of major objects from short videos. 
In particular, towards the high efficiency demands in practical short video application, a Temporal Information Fusion Network (TIFN) is proposed in the object detection module, which shows comparable accuracy and improved time efficiency to the state-of-the-art video object detector. 
Furthermore, in order to mitigate the fragmented issue of tracklets in short videos, a Cross-Layer Pointwise Siamese Network (CPSN) is proposed in the tracking module to enhance the robustness of the appearance model. 
Moreover, in order to evaluate the proposed system, two challenge datasets containing real-world short videos are built for video object trajectory extraction and generic object re-identification respectively. 
Overall, extensive experiments for each module and the whole system demonstrate the effectiveness and efficiency of our system.
\end{abstract}




%   \centering
%   \includegraphics[scale=0.22]{images/sys_framework.png}
%   \caption{Overview of the overall flowchart of the generic object re-identification system.}
%   \label{fig:system}

% \begin{figure*}[ht]
% \centering
% \includegraphics[scale=0.22]{images/sys_framework.png}
% %  \vskip -0.1in
% \caption{Overview of the overall flowchart of the generic object re-identification system.}
% %  \vskip -0.15in
% \label{fig:system}
% \end{figure*}

\maketitle

\begin{figure*}
  \centering
  \includegraphics[scale=0.55]{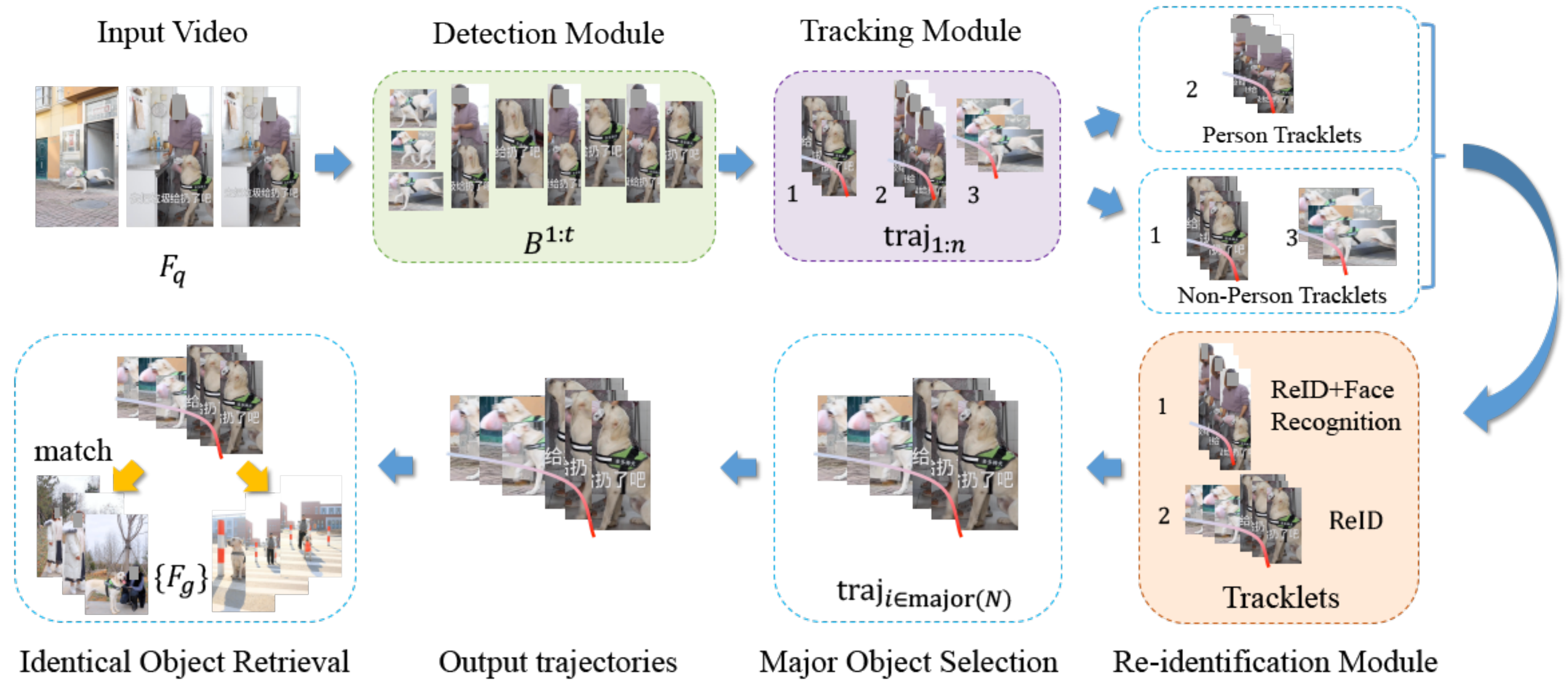}
  \caption{Overview of the overall flowchart of the generic object re-identification system.}
  \label{fig:system}
\end{figure*}

\section{Introduction}
\label{sec:intro}
Short video has been one of the most important social media and entertaining fashion for people worldwide recently. As millions of short videos are uploaded to the platforms by users each day, requirement of generic object re-identification from a large amount of short videos increases rapidly. 

However, building a generic object re-identification system for short videos poses a number of challenges due to characteristics of short videos as follows:

\begin{itemize}
 \item \textbf{Great variety in pattern and object category:} Most short videos in short video platforms are uploaded by users. Topics of video vary a lot according to interest of users, and numerous categories of objects may occur in the same video simultaneously.
 
 \item \textbf{Various visual effects (VFX):} Visual effects are common when different video shots switching or emphasizing objects in short videos. These visual effects may cause color change of video frame, object distortion, etc.
 
 \item \textbf{Complicated motions and postures:} Different from scenarios such as traffic monitoring, motion and posture of people in short videos are much more unpredictable. Moreover, frequent strenuous motion may also bring great burden to object detection and tracking.
 
 \item \textbf{Rapid changes in object appearance:} Different from traffic scenarios in most person re-identification tasks, people dress differently in different short videos. Even more, some short videos are consist of multiple video clips (or shots), and appearance of the same object changes rapidly when the video shot switches, such as in dress switching videos.
\end{itemize}
As a result, short video has become one of the most challenging scenarios in recent years.

To extract the trajectories of objects in videos, there exist a series of studies on video object detection and video object tracking. However, most of the public benchmark datasets only focus on relatively few categories. For example, ImageNet VID \cite{ILSVRC15}, which is most frequently used for evaluation in video object detection, only contains 30 categories, most of which are animals and vehicles. By contrast, there are hundreds of various categories, complicated visual effects (VFX) and multiple video shots in short video scenario, which is opposite to the scenario that most recent approaches assumed. 

In order to satisfy the mentioned challenge of generic object re-identification in short video scenarios, a novel system composed of a detection module, a tracking module, and a generic object re-identification module is proposed in this paper. In this framework, the trajectories of the major objects in the input query videos and their features are extracted by the three modules in the proposed system. As follow-up applications, the processed features can be applied in short video identical object retrieval tasks. The flowchart of the proposed system is illustrated in Figure \ref{fig:system}.

In particular, in order to handle the video object detection task with hundreds of categories, we build our detection module based on an image object detector, and propose a Temporal Information Fusion Network (TIFN) to take full advantage of temporal information among continuous frames. Further, a Cross-Layer Pointwise Siamese Network (CPSN) is proposed in the tracking module to mitigate the fragmented issue of tracklets, enhancing the robustness to match the same target despite the variety in appearance of objects in short videos.

To evaluate the proposed system, two challenge datasets are built for short video object trajectory extraction and generic object re-identification respectively. Extensive experiments for proposed methods and the whole proposed system are conducted with both the proposed short video datasets and public benchmark datasets, and the results demonstrate the high effectiveness and efficiency of the proposed system.

The contributions of this paper are summarized as follows:

\begin{itemize}
\item We design a generic object re-identification system for short videos. To the best of our knowledge, it is the first work to tackle the generic object re-identification problem under short video scenario, which is much more complicated and unpredictable in variety than regular benchmarks.

\item We propose TIFN, which takes full advantage of temporal information in videos to improve image object detector, and shows comparable accuracy and improved time efficiency to the state-of-the-art video object detector.

\item We propose CPSN, which makes full use of the local features in different scales, and excludes the disturb of the background and rapidly changing regions of foreground. This module can thus effectively reduce the fragmented issue of tracklets.

\item Two challenging datasets for short video object trajectory extraction and generic object re-identification are built, which have many potential applications in follow-up study of short video.
\end{itemize}

The rest of the paper is organized as follows. Section 2 discusses related work. The proposed framework and methods are detailedly described in Section 3. Proposed datasets introduction and experimental evaluations are shown in Section 4. Finally, Section 5 concludes this paper.

% \begin{itemize}
% \item[(1)] We design a novel system to handle generic object re-identification for short videos. To the best of our knowledge, it is the first work to tackle the generic object re-identification problem under short video scenario, which is much more complicated and unpredictable in variety than regular benchmarks.
% \item[(2)] We propose TIFN, which takes full advantage of temporal information in videos to improve image object detector, and shows comparable accuracy and improved time efficiency to the state-of-the-art video object detector.
% \item[(3)] We propose CPSN, which makes full use of the local features in different scales, and excludes the disturb of the background and rapidly changing regions of foreground. This module can thus effectively reduce the fragmented issue of tracklets.
% \item[(4)] Two challenging datasets for video object tracking and generic object re-identification for short videos are built, which have many potential applications in follow-up study of short video.
% \end{itemize}

\section{Related Work}
Building a system to handle generic object re-identification task among different short videos includes three primary sections: a video object detection module to locate objects, a video object tracking module to extract object tracklets in each shot, and a generic object re-identification module to retrieve tracklets of the same object in different shots to a complete trajectory. Therefore, the related work of the proposed can be concluded as follows.

{\bf Video Object Detection:} Video object detection (VOD) focuses on temporal information between adjacent frames. Combining detection and tracking, \cite{kang2017t} propose a multi-phase framework, including image object detection, bounding box tracking and temporal convolutional network to re-scoring tubelet, to obtain enhanced performance. \cite{luo2019detect} propose a scheduler network, which determines to detect or track at a certain frame. Without aid of tracking module, \cite{zhu2017flow} improve the per-frame features by aggregation of nearby features along the motion paths obtained by optical flow. Innovatively, \cite{wang2018fully} jointly calibrate the features of objects on both pixel-level and instance-level by optical flow, capturing detailed motion and global motion features. Instead of the time-consuming optical flow, \cite{Lu_2017_ICCV} use LSTM to extract temporal motion information. \cite{xiao2018video} propose a novel RNN architecture called the Spatial-Temporal Memory Network (STMM) to model both changing appearance and motion of objects over time. Considering the full-sequence level features, \cite{wu2019sequence} devise Sequence Level Semantics Aggregation (SELSA) module to obtain more discriminative and robust features for video object detection. \cite{chen2020memory} take full consideration of both global and local information and enlarge range of accessible content of the key frame by introducing a Long Range Memory (LRM) module. However, all the methods mentioned above either associate temporal information in frames by optical flow and RNN, or formed as a complicated and time-consuming framework. Moreover, all of them need training dataset where bounding boxes in every frame are well-annotated, which is difficult for labelling when categories increase. In contrast, the Temporal Information Fusion Network (TIFN) we propose is based on fast one-stage detector without extra supervision, so it can be trained with dataset for image object detection. 

%\subsection{Video Object Tracking}
% \noindent Recently most multiple object tracking methods focus on tracking the single category, generally pedestrians, because videos of pedestrians arise in a large number of practical applications. Additionally, much importance is attached to the surveillance scenarios. A standard tracking system can be divided into four stages: Detection stage, this stage is completed by the off-the-shelf detector. however, Tracktor creates a new tracking paradigm, a detector solve the most of tracking problems. Feature extraction stage, the DeepSort incorporated visual information extracted by a custom residual netowork, and proviedes a normalized vector with 128 features as output. Affinity stage, DAN learns compact, and comprehensive features of pre-detected objects at several levels of abstraction and compute the affinity matrix in an end-to-end manner. 

% However, methods mentioned above can only handle the person tracking under the traffic surveillance scenarios. In this paper , we propose an tracking system to adapt to generic object tracking under short video. 
{\bf Video Object Tracking:} Most of the recent multiple object tracking methods only focus on tracking single category, such as pedestrians or vehicles.
%because videos of pedestrians arise in a large number of practical applications. Furthermore, much importance is attached to the surveillance scenarios. 
A standard tracking system can be divided into four stages: detection stage, feature extraction stage, affinity stage and association stage, while recent works pay more attention to the second and third stages. DeepSort \cite{wojke2017simple} integrates the deep neural network to learn appearance information, and obtains improved performance compared with previously published approach SORT \cite{bewley2016simple}. DAN \cite{sun2019deep} combines the feature extraction stage and affinity stage together to learn compact, yet comprehensive features in an end-to-end manner. Unlike most methods which use the off-the-shelf detector to detect region of interest, Tracktor \cite{bergmann2019tracking} creates a new tracking paradigm, in which a single detector solve most of tracking problems. 
Unfortunately, most of these methods can only handle the pedestrian tracking problems under the traffic surveillance scenarios, which are not applicable to generic objects. In this paper, we propose a tracking system to adapt to generic object tracking for short videos.

%\subsection{Re-identification}
{\bf Re-identification:} Recently most re-identification (ReID) methods focus on person retrieval. A standard person re-identification system contains three main components: feature representation learning, deep metric learning and ranking optimization. Feature representation in ReID can be furtherly divided into global feature and local feature. Due to the complicated scenario in person ReID, local features such as partial region feature, human pose and landmark feature, are becoming principal components, rather than global features. \cite{zhang2017alignedreid} extract global feature which is jointly learned with local partial region features to enhance global feature learning. \cite{Zhao_2017_CVPR} employ human pose estimation and landmark detection in person re-identification. \cite{sun2018beyond} target in learning discriminative part-informed features, and propose Part-based Convolutional Baseline (PCB) and refined part pooling (RPP), resulting in refined parts with enhanced within-part consistency.

However, most methods mentioned above are not designed for short video scenarios. Due to the challenging characteristics of short videos described in Section \ref{sec:intro}, existing methods are difficult to be applied to generic object re-identification for short videos directly. Considering numerous categories, various VFX, multiple video shots and rapid appearance changes, we propose a novel system to handle generic object re-identification for short videos in this paper. To our knowledge, it is the first system to tackle generic object re-identification problem for short video, which is one of the most challenge task in recent years.

\section{Proposed Method}
\noindent As is shown in Figure \ref{fig:system}, a generic object re-identification system for short videos can be formulated as follows. Given a sequence of video frames $F_q = \{f_t\}_{t=1}^T$ which contains $N$ objects and $K$ shots, the goal is to obtain the trajectories of the major objects $\{{\rm traj}_n\}_{n \in {\rm major}(N)}$ and use their features $\{{\rm feat}_n\}_{n \in {\rm major}(N)}$ to retrieve other videos $\{F_g\}$ with identical objects in gallery, where ${\rm major}(N)$ denotes the major objects of the query video. Shots in the query video are denoted as $S = \{s_k\}_{k=1}^K$, where $s_k = \{f_t\}_{t=t^*}^{t^*+{\rm length}(s_k)-1}$ denotes the $k$th shot in video which started at $t=t^*$. 

Considering the complex scenario in short video and robustness of algorithm, we decompose the generic object re-identification for short videos problem into several sub-problems, and formulate them as an end-to-end system, which consists of three major modules: detection module, tracking module and re-identification module (colored in Figure \ref{fig:system}). An algorithm summary follows the introduction of the three modules.

% We first detect bounding boxes $B^t=\{b_n^t\}_{n=1}^N$ of the $N$ objects in each frame $f_t$. Then all bounding boxes of the $n$th object in the $k$th shot are connected to generate tracklet $tr_{nk} = \{b_n^t\}_{t=i}^{i+length(s_k)-1}$. Next, tracklets of the $n$th object from $K$ shot are grouped into a complete trajectory $traj_n = \{tr_{nk}\}_{k=1}^K$, and features of objects per frame in trajectory $traj_n$ are sampled and averaged as the features of trajectory $traj_n$, denoted as $feat_n$. At last, major trajectories $\{traj_n\}_{n=major(N)}$ are selected by duration time and mean size, and features of major objects $\{feat_n\}_{n=major(N)}$ are selected relatively. A similarity score is computed between each pair of major object features in query video and gallery video to evaluate whether existing identical object or not.

\subsection{Detection Module}
\noindent In order to extract feature embedding of each object in video accurately, all the candidate objects need to be localized at first. The aim of detection module is to detect bounding boxes $B^t=\{b_n^t\}_{n=1}^N$ of the $N$ objects in each frame $f_t$. 

\begin{figure}[t]
\centering
\includegraphics[width=\linewidth]{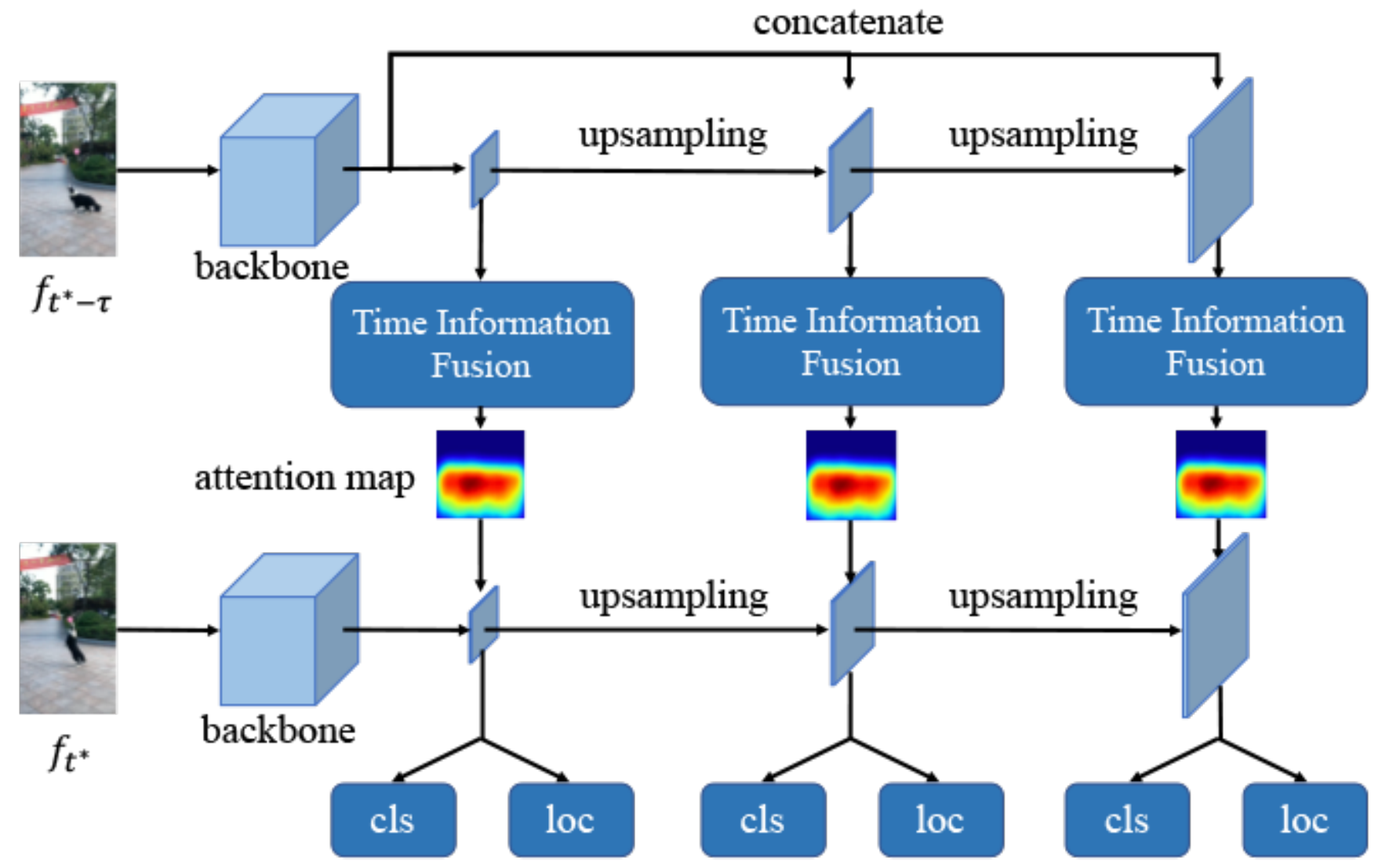}
 \vskip -0.1in
\caption{\small Visualization of the architecture of the proposed TIFN. The details of the $3$ structure striding over the time are shown in Figure \ref{fig:net_tifn_sub}.}
 \vskip -0.1in
\label{fig:net_tifn}
\end{figure}

\begin{figure}[t]
\centering
\subfigure[]{
    \centering
    \includegraphics[width=0.9\linewidth]{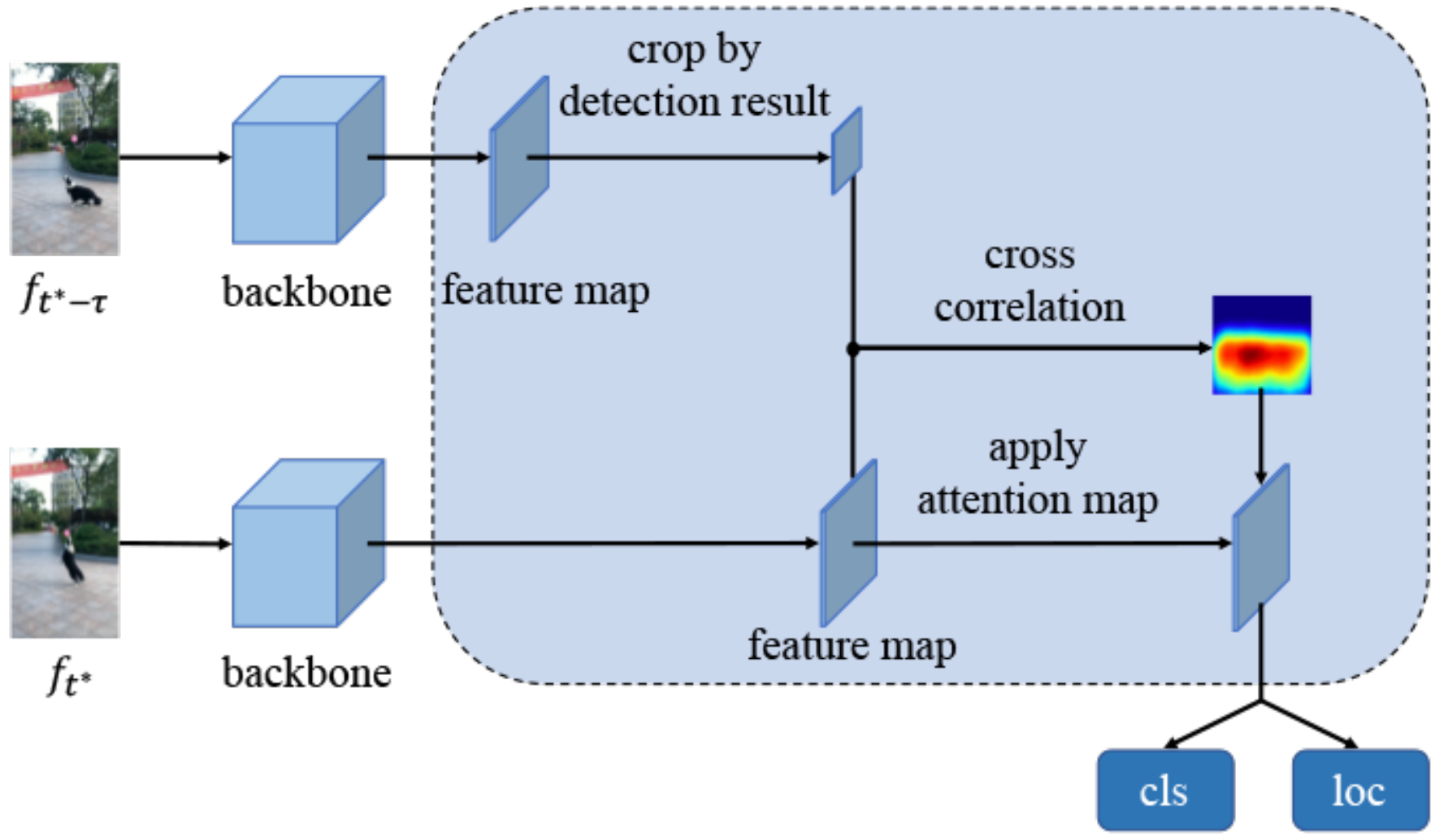}
    \label{subfig:tifn_ls}
}
\subfigure[]{
    \centering
    \includegraphics[width=0.9\linewidth]{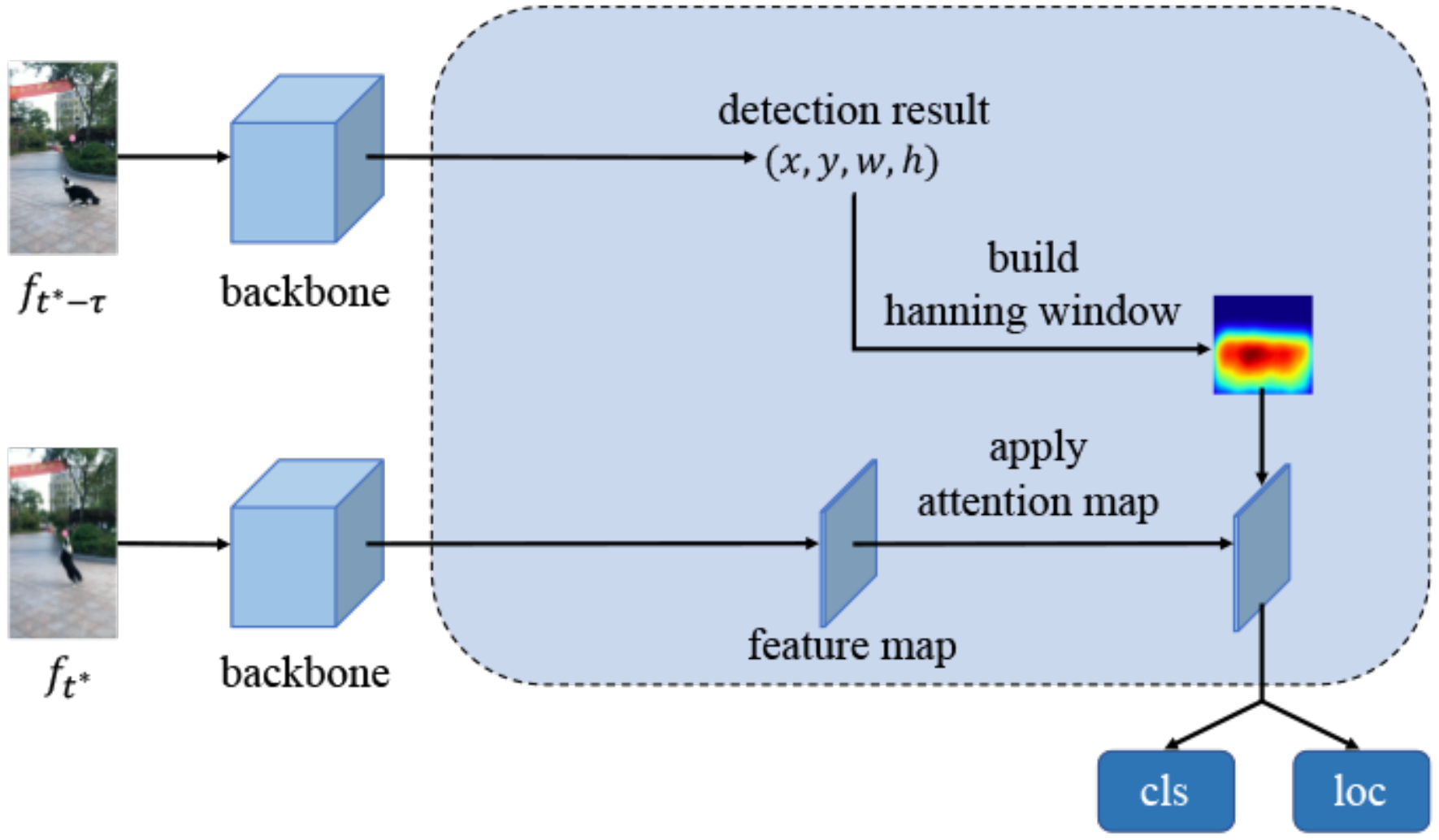}
    \label{subfig:tifn_ll}
}
\subfigure[]{
    \centering
    \includegraphics[width=0.9\linewidth]{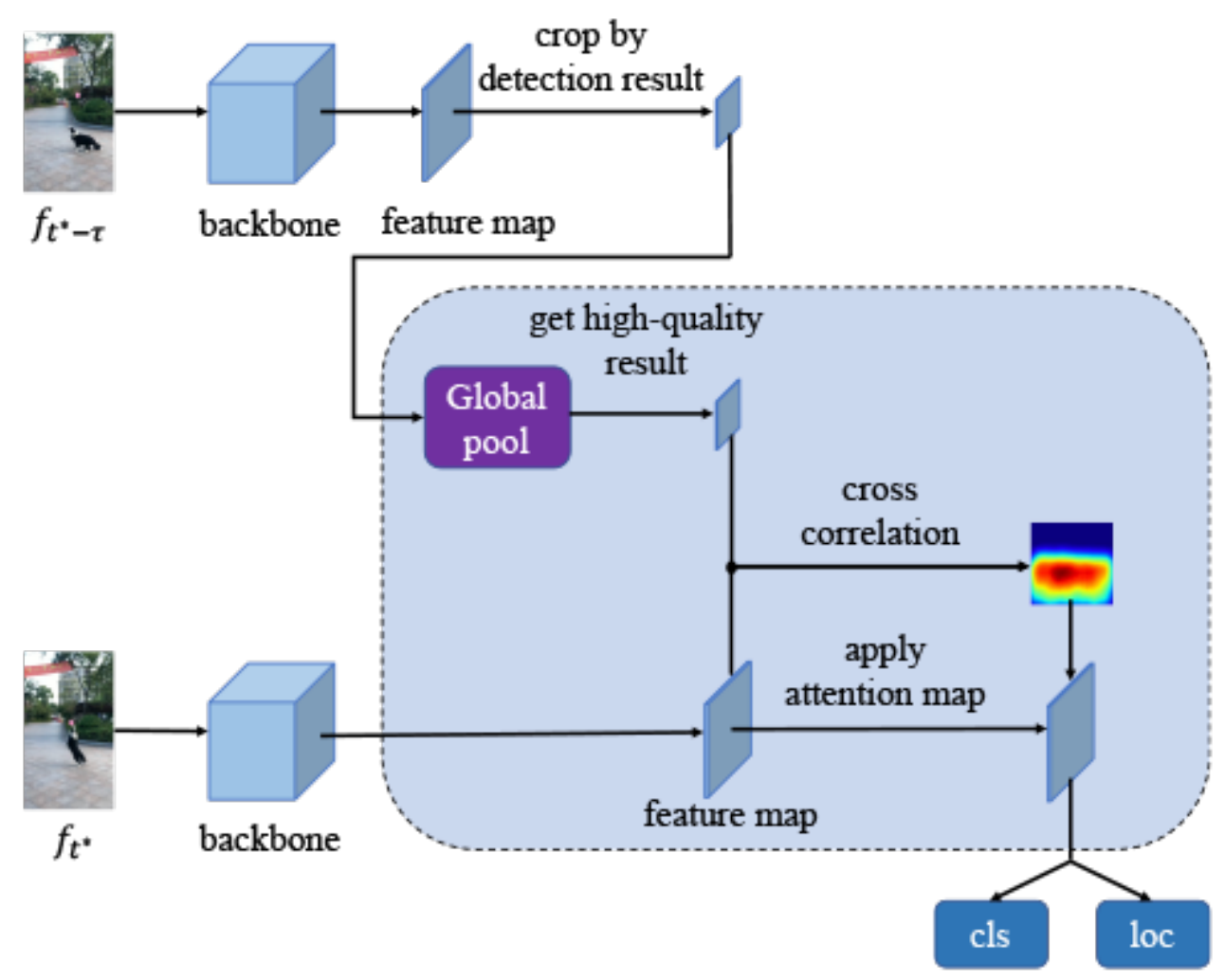}
    \label{subfig:tifn_gs}
}
\caption{\small \subref{subfig:tifn_ls} is the local semantic information fusion module, \subref{subfig:tifn_ll} is the local location information fusion module and \subref{subfig:tifn_gs} is the global semantic information fusion module.}
\label{fig:net_tifn_sub}
\end{figure}

To detect objects in video, most works utilize optical flow motion or on the basis of two-stage detector like Faster-RCNN \cite{ren2015faster}, which are time consuming. Moreover, to train these models, dataset that objects are well-annotated in every frame is required. When category of object increases, the time cost of labelling all frames in video increases proportionally as well. In view of the great variety of categories in short video and the efficiency requirements in practical applications, we start from a fast one-stage detector YOLOv3-SPP \cite{redmon2018yolov3, he2015spatial} for single image detection, and take full advantage of temporal information in video to improve the accuracy by introducing the proposed Temporal Information Fusion Network (TIFN) without extra supervision.

Inspired by \cite{chen2020memory}, the temporal information in video can be summarized as three parts: local semantic information, local localization information and global semantic information. Local semantic information is the appearance similarity of the same object between adjacent frames, and local localization information represents the continuity of spatial information (i.e., height, width and center location) of the identical object in continuous frames. Global semantic information, which is similar to the local one, means the appearance similarity of the same object in the entire video, but due to practical implementation and efficiency, we only consider frames before the frame at current $t^*$. Figure \ref{fig:net_tifn} is an illustration of the detection module, and Figure \ref{fig:net_tifn_sub} shows the three time information fusion modules.

According to the candidate boxes predicted in each frame, the corresponding regions in the three feature maps before each detection head of three scales are cropped and saved with their spatial information, i.e., center position, width and height, denoted as $\{{\rm crop}_n^t\}_{t=1, n=1}^{T, N}$ and $\{{\rm spat}_n^t\}_{t=1, n=1}^{T, N}$. 

For local semantic information, as illustrated in Figure \ref{fig:net_tifn_sub}\subref{subfig:tifn_ls}, when detecting objects at $t=t^*$, cross-correlation is calculated between each cropped feature map  $\{{\rm crop}_n^t\}_{t=t^*-\tau, n=1}^{t^*-1, N}$ in the past $\tau$ frames and feature map of $f_{t^*}$
{\small\begin{equation}
\label{con:attn_ls}
    {\rm Attn_{ls}}(f_{t^*}) = \frac{1}{\tau} \sum_{t=t^*-\tau}^{t^*-1}{\rm clip} \Big( {\sum_{n=1}^N{{\rm xcorr} \big( {\rm crop}_n^t,  \Phi(f_{t^*}) \big) }} \Big)
\end{equation}
}where ${\rm Attn_{ls}}$ denotes the attention map of $f_{t^*}$ obtained by local semantic information, i.e., the mean cross-correlation heatmap obtained, ${\rm clip}(\cdot)$ denotes clipping value to $[0, 1]$, ${\rm xcorr}(\cdot)$ denotes calculating cross-correlation, and $\Phi(\cdot)$ denotes the neural network for feature extraction in detector.

For local localization information, as illustrated in Figure \ref{fig:net_tifn_sub}\subref{subfig:tifn_ll}, attention map is obtained simply by applying truncated 2-Dimension hanning window at the corresponding coordinates to the saved center position
{\small\begin{equation}
\label{con:attn_ll}
    {\rm Attn_{ll}}(f_{t^*}) = \frac{1}{\tau}\sum_{t=t^*-\tau}^{t^*-1} \sum_{n=1}^N {\rm hanning}({\rm spat}_n^t)
\end{equation}
}where ${\rm Attn_{ll}}$ denotes the attention map of $f_{t^*}$ obtained by local localization information and ${\rm hanning}(\cdot)$ denotes applying truncated 2D hanning window with threshold $\lambda$.

For global semantic information, as illustrated in Figure \ref{fig:net_tifn_sub}\subref{subfig:tifn_gs}, the key is to utilize high-quality frames in global time domain of the input video to improve detection of local low-quality frames. Since the appearance and position of objects in short video may change rapidly and with low-quality due to the frequent video shot switching or rapid motion, local semantic information and local location information are not sufficient enough. However, high-quality frames in global with similar semantic information can be used to lead the attention to focus on objects in these low-quality frames.

To capture the global feature of objects, we build a global set $G$ with fixed size $\alpha_1$ which stores features of objects with high frequency. Similar to local semantic information, cropped features in each frame are obtained, and firstly added to a candidate pool $C$ with fixes size $\alpha_2$ ($\alpha_2 > \alpha_1$). Features in the candidate pool are matched with cropped feature of each bounding box in the next frame by similarity and updated according to the strategy as follows:
\begin{itemize}
    \item if not matched and $C$ is not full, add the new feature to $C$ directly.
    \item if not matched, $C$ is full and confidence of the new feature is higher than the lowest confidence in pool, replace the feature with lowest confidence.
    \item if matched, recorded frequency of matched feature adds 1, and if confidence of the new feature is higher than which matched, replace it.
\end{itemize}

When the recorded frequency of features in $C$ exceed the threshold $\gamma$, features are added to the global set $G$ if $G$ is not full. If $G$ is full, the feature with lowest confidence in $G$ is replaced only if the feature in $C$ is with higher confidence and frequency. Then cross-correlation map is obtained between each cropped feature map ${\rm crop}_n^t \in G$ and feature map of $f_{t^*}$, which is similar to the local semantic information.
{\small\begin{equation}
\label{con:attn_gs}
    {\rm Attn_{gs}}(f_{t^*}) = {\rm clip} \Big( \sum_{{\rm crop}_n \in G}{{\rm xcorr} \big( {\rm crop}_n,  \mathcal{N}(f_{t^*}) \big) } \Big)
\end{equation}
}where ${\rm Attn_{gs}}(f_{t^*})$ denotes the attention map of $f_{t^*}$ obtained by global semantic information. All the formulations mentioned above are with no extra parameters to be trained and can be done by processing each video frame only once.

At last, $3$ groups of candidate boxes are predicted by the feature maps applied with Equation (\ref{con:attn_ls})-(\ref{con:attn_gs}) respectively, and proceeded to post-processing phase such as Non-Maximum Suppression (NMS) together to obtain the final predicted bounding boxes $B^t$.

\subsection{Tracking Module}
\noindent After the detection stage, we could obtain a series of bounding boxes $B^t=\{b_n^t\}_{n=1}^N$. The aim of the tracking module is to link each bounding box of the same object. So we have to extract the features of those bounding boxes, compute the affinity matrix between detection responses and tracklets, and assign the matched bounding boxes to the existing tracklet frame-by-frame. The paper in this part focus on the feature extraction stage. Most existing methods focus on the global features, i.e., image-level features. \cite{li2019revisiting} argues that a measure in such a level may not be effective enough in light of the scarcity of examples under our scenario. Inspired by the  paper, we could obtain a more fine-grained feature based on the local features. 

\begin{figure}[t]
\centering
\includegraphics[width=\linewidth]{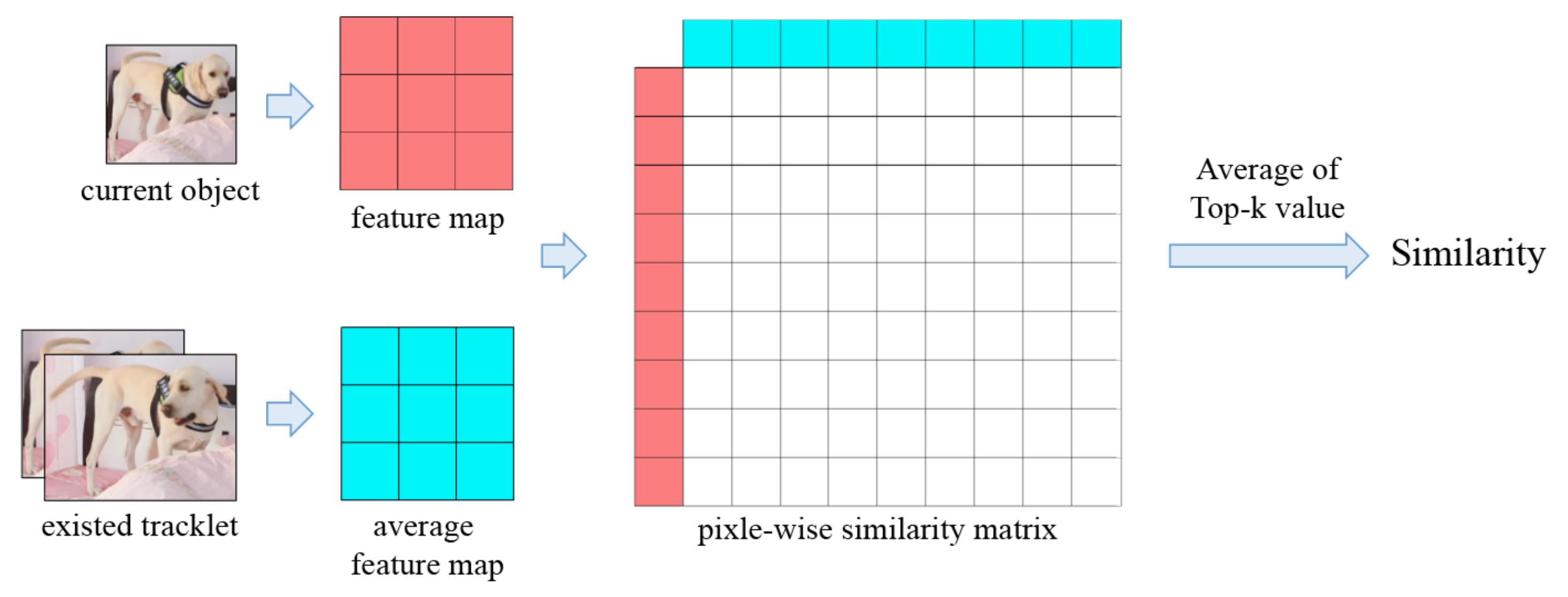}
\caption{\small Visualization of the architecture of the proposed CPSN.}
\label{fig:net_cpsn}
\end{figure}

In order to be consistent with the feature extractor in the following re-identification module, we first employ ResNet50 \cite{he2016deep} to learn the feature map with spatial feature. Given the cropped feature map ${\rm crop}_n \in C \times H_1 \times W_1$ derived by the detected bounding box $b_n$ ($t$ is ignored for convenience of description) and a single ${\rm crop}_d \in C \times H_2 \times W_2$ from existing tracklet, we try to obtain a pointwise response map. 

$x_{ij}$ denotes the cell in feature map, i.e., the local feature of the target. For convenience of calculation, cosine similarity metric is used to measure the similarity of the local region between detection response and tracklet. Naturally, the pointwise similarity could be formulated as follows:
{\small\begin{equation}
\label{con:simi}
    {\rm sim}(x^d_{ij}, x^t_{i'j'})=\cos(x^d_{ij}, x^t_{i'j'})
\end{equation}
}
{\small\begin{equation}
\label{con:simi2}
    A^s=x^{d^{\mathsf{T}}}x^t
\end{equation}
}where ${\rm sim}(\cdot)$ denotes the similarity between different cells and $A^s$ denotes the pointwise response map. We can just use matrix multiplication to calculate the response map, and the amplitude of each cell of it represents the similarity of the part from the pair. So we could use the mean value of the top$k$ maximum response to approximate the original similarity of the pair. This score could exclude the disturb of the background and rapidly changing parts of foreground. 

The operation above just happens on one stage. Under short-video scenarios, there always be scale changes, so the same target may occupy in different scales in different two frames. As a result, we use multi-scale feature of region in two frames to find the regions matched with highest response. Briefly, $x_1^d, x_2^d, \cdots, x_n^d$ denote the feature map in different scales of the detection response, and similarly, $x_1^t, x_2^t, \cdots, x_n^t$ denote the feature map in different scales of the tracklet. So we could integrate multiple layers in different scale and averaging the score mentioned above, and get the final similarity score. 

{\small\begin{equation}
    % {\rm SIM}(x^d,x^t)={\rm mean}\{{\rm mean}\{{\rm top} k\{A^s_{mn}\}\}\}_{m=1, n=1}^{N, N}
    {\rm SIM}(x^d,x^t)=\frac 1 {N^2} \sum\limits_m\sum\limits_n\{{\rm top} k\{A^s_{mn}\}\}_{m=1, n=1}^{N, N}
\end{equation}
}where ${\rm top} k(\cdot)$ denotes the top$k$ values of cells in the matrix $A^s$, and ${\rm SIM}(\cdot)$ denotes the similarity between matrices, which is different from the similarity between cells in Equation (\ref{con:simi}).

For other parts such as motion prediction branch, bipartite matching, trajectory management, we just keep the same as DeepSort, due to its high efficiency. 

\subsection{Re-Identification Module}
\noindent In re-identification module, tracklets of the $n$th object from $K$ shots are grouped into a complete trajectory $traj_n = \{{\rm tr}_{nk}\}_{k=1}^K$, and features of objects per frame in trajectory ${\rm traj}_n$ are sampled and averaged as the features of trajectory ${\rm traj}_n$, denoted as ${\rm feat}_n$. Then major trajectories $\{{\rm traj}_n\}_{n \in {\rm major}(N)}$ are selected by duration time and mean size, and features of major objects $\{{\rm feat}_n\}_{n \in {\rm major}(N)}$ are selected relatively. A similarity score is computed between each pair of major object features in query video and gallery video to retrieve videos with identical objects.

In retrieving objects of general categories for short videos, for person, we need to identify each instance since they are different in appearance. But for non-person object, such as animal or commodity, instances of same breed or same pattern are unrecognizable in appearance mostly. As a result, the re-identification module is divided to two branches: person branch and non-person object branch, dealing with each kind of object respectively.

In the person branch, we apply the person re-identification process with face detection and recognition, since in different short videos, the same person may dress in different clothes. As a result, a face detection and recognition model is introduced to support person re-identification. We use the PCB network \cite{sun2018beyond} as the person ReID feature extractor and ArcFace \cite{deng2019arcface} as the face feature extractor. The person similarity is decided by cosine similarity computed between each pair of features obtained by the two models together as follows

{\small\begin{equation}
    S_{\rm person} = \left\{
    \begin{array}{ll}
        \lambda_1 \times S_{\rm ReID} + \lambda_2 \times S_{\rm Face}   &,\ {\rm face}\ {\rm detected}       \\
        S_{\rm ReID}                                                    &,\ {\rm not}\ {\rm detected}
    \end{array} \right.
\label{con:simi_person}
\end{equation}}

In the non-person object branch, since we regard unrecognizable objects of same breed or same pattern as same instance (unless they appear simultaneously), we formulate the non-person object re-identification as a fine-grained classification problem. The first four blocks of ResNet50 \cite{he2016deep} trained by ImageNet dataset \cite{deng2009imagenet} with a average pooling layer followed are applied in non-person object branch as feature extractor. ImageNet dataset consists of $1000$ fine-grained categories, as a result of which, discriminative feature can be obtained.

Furthermore, since general re-identification architectures retain a fixed query set, tracklets appeared in the first frame in most cases, to retrieve identical objects in the gallery set, which is hard to satisfy the complicated variety of appearance and pose in short videos and is liable to cause mismatching, an updating mechanism is introduced to query set in both person branch and non-person object branch in re-identification module. Setting tracklets appeared in the first frame as query set similarly, when each tracklet remained in gallery set is accessed individually, similarities between query tracklets and it are first computed. Contrast to general formulation, the accessed tracklet is updated to the query set if the difference between mean similarity among matched tracklets and mean similarity among unmatched tracklets is larger than a threshold $\Delta$. Therefore, the proposed updating mechanism can mitigate mismatching caused by complicated variety of appearance and pose in short videos.

In identical object retrieval among different videos, first the major objects are selected according to the length and the average size of predicted trajectories. Then the average features of each major object are calculated by re-identification module and used to match other features of major objects in the short video gallery according to cosine similarity metrics.

\section{Experiments}
\noindent To demonstrate the performance of the proposed generic object re-identification system for short videos, two novel datasets for evaluation are proposed in this paper, which comprised of various short videos collected from short video platform. Besides, the proposed generic object re-identification system is compared with other approaches under public benchmark, such as ImageNet VID \cite{ILSVRC15} and MOTChallenge \cite{PoseTrack}, to indicate the generalization ability.

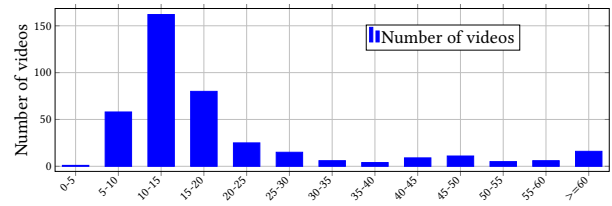
\begin{figure}[t]
\begin{tikzpicture}[scale=0.40]
\begin{axis}[
height = 7cm,
width = 20cm,
grid=major,
x tick label style = {rotate=45,anchor=east,font=\LARGE},
y tick label style = {font=\LARGE},
ylabel style = {font=\Huge,yshift=0pt},
ylabel=Number of videos,
%xlabel=Labels,
%enlarge x limits =0.025pt,
enlarge x limits =0.04,
enlarge y limits =0.04,
legend style={area legend,at={(0.7,0.9),font=\Huge},anchor=north, legend columns=-1},
ybar=0.45pt,% configures `bar shift'
bar width=25pt,
%ybar interval = 0.7pt,
%nodes near coords,
point meta=y,% the displayed number
%xticklabels={,person,dog,car,chair,cat,bird,bottle,aeroplane,sofa,diningtable,tvmonitor,pottedplant,bicycle,motorbike,train,boat,horse,bus,sheep,cow},
%]
xticklabels={,0-5,5-10,10-15,15-20,20-25,25-30,30-35,35-40,40-45,45-50,50-55,55-60,$>=$60},
]
\addplot[draw=blue,fill=blue]
coordinates {(0,1)(1,58)(2,162)(3,80)(4,25)(5,15)(6,6)(7,4)(8,9)(9,11)(10,5)(11,6)(12,16)};
%coordinates {(1,1)(2,58)(3,162)(4,80)(5,25)(6,15)(7,6)(8,4)(9,9)(10,11)(11,5)(12,6)(13,16)};
%\addplot[draw=red,fill=red]
%coordinates {(0,1.0000)(0.5,0.1231)(1.0,0.1943)(1.5,0.0746)(2,0.0846)(2.5,0.2232)(3,0)(3.5,0.4277)(4,0.1331)(4.5,0.0077)(5,0)(5.5,0.0308)(6,0.1024)(6.5,0.1432)(7,0.0846)(7.5,0.2594)(8,0.1685)(8.5,0.3465)(9,0.4071)(9.5,0.1967)};
\legend{Number of videos}
%set(gca,xticklabels,[0:0.5:9.5]);
%set(gcf,'resizeFcn','l=get(gca,''xlim''),a=get(gcf,''position'');d=560/a(3);
%set(gca,'xticklabels',round([l(1):75*d:l(2)]))');
\end{axis}
\end{tikzpicture}
\caption{\small Video length distribution of the proposed SVD-IOR dataset.}
 \vskip -0.15in
\label{fig:video_len}
%\vspace*{-0.5em}

\end{figure}

\begin{figure}[H]
\centering
\includegraphics[width=0.9\linewidth]{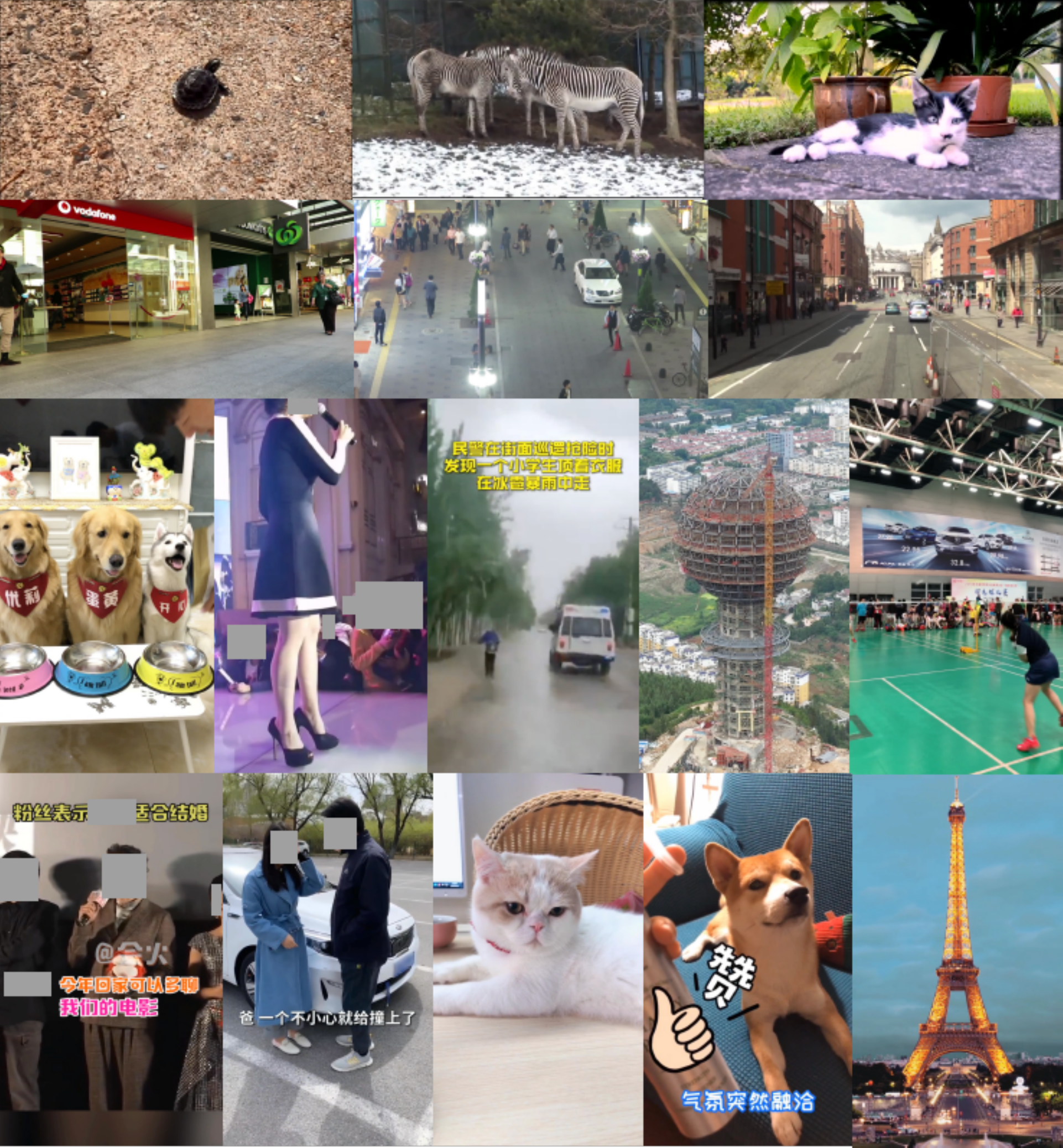}
\vskip -0.05in
\caption{Visualization of the comparison among ImageNet VID, MOTChallenge, SVD-IOR and SVD-ReID.}
\vskip -0.15in
\label{fig:diff_dataset}
\end{figure}

\subsection{Proposed Datasets}
To compare with existing methods and determine state-of-the-art approach under short-video scenarios, we propose two novel short video dataset for evaluation in identical object retrieval among different shots and different videos. Videos in the proposed datasets are both collected from real-world short video platform. 

The short video dataset for identical object retrieval among different shots, abbreviated as SVD-IOR, includes $400$ videos in amount with a variety of characteristics, e.g., multiple categories, complex scenarios and frequent shot changes. Foreground objects are annotated every $10$ frames in the form of bounding boxes with corresponding track ID in each video. As is shown in Figure \ref{fig:video_len}, the lengths of most videos are distributed between $10$ and $20$ seconds. In total, it contains $232964$ frames at various resolutions, with $1221$ different identities and $66180$ bounding boxes annotated. Comparisons among ImageNet VID, MOTChallenge and the proposed SVD-IOR, SVD-ReID are shown as Figure \ref{fig:diff_dataset}.

The short video dataset for generic object re-identification among different videos, abbreviated as SVD-ReID, contains $400$ short videos. Objects of four categories, i.e., person, cat, dog, landmark, are annotated as same as which in SVD-IOR. Videos of each class contain $10$ different object instances and each instance contains $10$ different videos collected from corresponding user. Specially, track ID of each identical object among different videos is annotated as same value. The $10$ videos of each object instance are split into $2$ query videos and $8$ gallery videos, and aggregated to build a query set and a gallery set in order to evaluate generic object re-identification among different videos.

To demonstrate the generalization ability of our method on public benchmarks, the proposed system is evaluated on benchmark dataset ImageNet VID and MOTChallenge respectively as well.

\subsection{Implementation Details}

\noindent For detection module during training, we follow the settings in \cite{redmon2018yolov3} to train YOLOv3-SPP model with training set of OpenImage \cite{OpenImages2} and ImageNet VID, respectively evaluating on the proposed short video dataset and ImageNet VID validation set. During evaluation, the input videos are first split by shots with shot boundary detection approach based on color histogram \cite{mas2003video} to ensure tracklets of objects are continuous in each shot. The input video frames are resized and padded to $608 \times 608$.

% Then, frames are resized and padded to $608 \times 608$, and the hyperparameter in Equation (\ref{con:attn_ls}-\ref{con:attn_gs}) is set as $\tau=3$. The confidence threshold of predicted bounding boxes is set as $0.2$ and the IoU threshold in NMS is set as $0.4$. Since local localization and global semantic information are particularly designed for varying appearance of obejcts in short video scenarios, only local semantic information is aggregated in detection module for ImageNet VID dataset. 

For tracking module, We use the training set of MOTChallenge to train, while using the validation set to evaluate. For the proposed CPSN, we rescale the image pair to $256\times 256$ as inputs during training. With ResNet50 as the backbone, outputs of the second and forth stage are used to build the response map $A^s$ in Equation \ref{con:simi2}. It is worth noting that the calculation of the response map don't introduce any extra parameters. Circle loss and Adam optimizer are employed to optimize the proposed CPSN. 
% Besides, we keep the other thresholds in consistent with DeepSort \cite{wojke2017simple}. 

% Since existing tracking methods focus on the pedestrian tracking under traffic surveillance scenarios, we have to modify the other methods to adapt to our scenarios for fair comparison. For Tracktor, due to the large-amplitude movements in short video, which contradict the assumption in Tracktor that targets are moving slightly between adjacent frames, we add a resampling mechanism to enlarge the searching region in the RPN stage. For DAN, due to the lack of detector as described in \cite{sun2019deep}, we use our video object detector instead. 

For re-identification module during training, we follow the settings in \cite{sun2018beyond} and train the PCB network as the person re-identification model with Market-1501 \cite{zheng2015scalable}, while using ResNet50 pretrained by ImageNet \cite{deng2009imagenet} as the non-person object re-identification model to extract fine-grained feature.

% The amount of stripes $p$ in PCB is set to $6$ as \cite{sun2018beyond} described, and coefficient parameters in Equation \ref{con:simi_person} are set as $\lambda_1=0.22$ and $\lambda_2=0.78$. Threshold $\Delta$ is set to $0.3$ in both person branch and non-person object branch.

\subsection{Evaluation of Detection Module}

For video object detection under short video dataset, ablation experiments with SVD-IOR are applied to demonstrate the effectiveness of the proposed TIFN. Moreover, performance comparison between the proposed TIFN and the state-of-the-art video object detection models on ImageNet VID validation set are reported. Quantitative comparison results are summarized in Table \ref{tab:det_ablation} and Table \ref{tab:det_compare}.

Table \ref{tab:det_ablation} describes the ablation experiment results of aggregation of local semantic information, local localization and global semantic information under short video dataset. As is shown in the table, the proposed TIFN obtains improvement of approximately $5\%$ mAP rather than the base model YOLOv3-SPP, without extra supervision. The introduction of local semantic information contributes most, since the temporal information between adjacent frames is most significant in videos. The improvement brought by local localization and global semantic information indicate their effectiveness in dealing with the rapid change of appearance in short videos. Besides, comparison results of execution efficiency show that the proposed TIFN can still keep high efficiency after aggregating multiple temporal information.

%%%%%%%%%%%%%%%%%%%%% 
 \newsavebox{\taba}
 \begin{lrbox}{\taba}
\begin{tabular}{c|ccc|c|c}
    \toprule[2pt]
    \multirow{2}{*}{Methods}& local     & local         & global    & \multirow{2}{*}{mAP(\%)}  & \multirow{2}{*}{FPS}  \\
                            & semantic  & localization  & semantic  &                           &                       \\
    \midrule[1pt]
    base model \cite{redmon2018yolov3, he2015spatial}              &           &               &           & 60.91                     & 15.27                 \\
    + ls                    &\checkmark &               &           & 64.73                     & 15.01                 \\
    + ls + ll               &\checkmark & \checkmark    &           & 65.06                     & 12.51                 \\
    + ls + gs               &\checkmark &               &\checkmark & 65.47                     & 10.79                 \\
    + ls + ll + gs          &\checkmark & \checkmark    &\checkmark & \textbf{65.69}            & 10.65                 \\
    \bottomrule[2pt]
\end{tabular}
 \end{lrbox}
%%%%%%%%%%%%%%%%%%%%%%%%%%

 %\begin{table}[htbp]
 \begin{table}[t]
 \caption{\small Ablation study on aggregation of local semantic information, local localization and global semantic information for short videos. } 
 \vskip -0.1in
 \centering   
      \scalebox{0.8}{\usebox{\taba}}
\label{tab:det_ablation}
 %\vskip -0.1in
 \end{table}  
%%%%%%%%%%%%%%%%%%%%%%%%%%%%%%%%%%%%%%%%%%%%%%%%%%%%%%

As is shown in Table \ref{tab:det_compare}, TIFN performs comparable accuracy and improved time efficiency to the state-of-the-art approaches. Since most methods shown in Table \ref{tab:det_compare} are formulated in much more complex structure with complicated post-processing, the proposed TIFN with a simple backbone is more suitable for real-world applications.

%%%%%%%%%%%%%%%%%%%%% 
 \newsavebox{\tabb}
 \begin{lrbox}{\tabb}
\begin{tabular}{c|c|c|c}
    \toprule[2pt]
    Methods         & Backbone          & mAP(\%)           & FPS    \\
    \midrule[1pt]
    T-CNN \cite{kang2017t}           & DeepID-Net+CRAFT \cite{ouyang2015deepid, baek2019character}  & 73.8              & ---    \\
    FGFA \cite{zhu2017flow}            & ResNet-101 \cite{he2016deep}        & 78.4              & 1.14   \\
    DoT \cite{luo2019detect}             & ResNet-101        & 79.8              & ---    \\
    MANet \cite{wang2018fully}           & ResNet-101        & 80.3              & 4.96   \\
    SELSA \cite{wu2019sequence}           & ResNet-101        & 80.5              & ---    \\
    MEGA \cite{chen2020memory}            & ResNet-101        & \textbf{82.9}     & 8.73   \\
    \midrule[1pt]
    FGFA \cite{zhu2017flow}            & Inception-ResNet \cite{szegedy2017inception}  & 80.1              & 1.05   \\
    DoT \cite{luo2019detect}             & Inception-v4 \cite{szegedy2017inception}      & 82.1              & ---    \\
    SELSA \cite{wu2019sequence}           & ResNeXt-101 \cite{xie2017aggregated}       & 84.3              & ---    \\
    MEGA \cite{chen2020memory}            & ResNeXt-101       & \textbf{85.4}     & 1.10   \\
    \midrule[1pt]
    base model \cite{redmon2018yolov3, he2015spatial} & DarkNet53    & 74.3              & \textbf{17.40}  \\
    TIFN (ours)     & DarkNet53         & \textbf{83.2}     & \textbf{10.14}  \\
    \bottomrule[2pt]
\end{tabular}
 \end{lrbox}
%%%%%%%%%%%%%%%%%%%%%%%%%%

 %\begin{table}[htbp]
 \begin{table}[H]
 \caption{\small Comparison results among the proposed TIFN and other video object detection methods under ImageNet VID dataset. } 
 \vskip -0.1in
 \centering   
      \scalebox{0.85}{\usebox{\tabb}}
\label{tab:det_compare}
 \vskip -0.1in
 \end{table}  
%%%%%%%%%%%%%%%%%%%%%%%%%%%%%%%%%%%%%%%%%%%%%%%%%%%%%%

% \begin{table}[htbp]
% \centering
% \begin{tabular}{c|c|c|c|c}
%     \toprule[2pt]
%     \diagbox{Dataset}{MOTA(\%)}{Methods}    & Tracktor  & DAN   & ours              \\
%     \midrule[1pt]
%     PoseTrack                               & 37.58     & 41.69 & \textbf{43.26}    \\
%     VidOR                                   & 21.58     & 20.69 & \textbf{37.36}    \\
%     Short video dataset (ours)              & 36.98     & 33.79 & \textbf{41.85}    \\
%     \bottomrule[2pt]
% \end{tabular}
% \caption{Comparison results among the proposed system, Tracktor and DAN under PoseTrack, VidOR and the proposed short video dataset.}
% \label{tab:track_compare}
% \end{table}

\subsection{Evaluation of Tracking Module}

For evalution of tracking module, the proposed CPSN is compared with numerous classic and recent object tracking methods both benchmark dataset MOTChallenge and the proposed short video dataset SVD-IOR.

As is shown in Table \ref{tab:track_compare_motchallenge}, the proposed multi-object tracking method in this paper performs poor results on the public datasets, since it is not designed for the surveillance scenarios in MOTChallenge dataset. The better performance than jCC and FAMNet demonstrate the great generalization of the proposed CPSN to some extent.

%%%%%%%%%%%%%%%%%%%%%% cgx %%%%%%%%%%%%%%%%%%%%%%%%%%
\newsavebox{\tabc}
\begin{lrbox}{\tabc}
\begin{tabular}{c|c|c|c|c}
    \toprule[2pt]
    \diagbox{Methods}{Metrics}      & MOTA      & FP        & FN        & IDS       \\
    \midrule[1pt]
    Tracktor++v2 \cite{bergmann2019tracking}                   & 56.3      & \textbf{8866}      & 235449    & 1987      \\
    GSM\_Tracktor \cite{liugsm}                  & 56.4      & 14379     & 231074    & 1485      \\
    Lif\_TsimInt \cite{hornakova2020lifted}                    & 58.2      & 16850     & 217944    & \textbf{1022}      \\
    MPNTrack \cite{braso2020learning}                        & 58.8	    & 17413     & 213594    & 1185      \\
    Lif\_T \cite{hornakova2020lifted}                          & 60.5      & 14966     & 206619    & 1189      \\
    CTTrackPub \cite{zhou2020tracking}                      & \textbf{61.5}      & 14076     & \textbf{200672}    & 2583      \\
    jCC \cite{keuper2018motion}                             & 51.2      & 25937     & 247822    & 1802      \\
    FAMNet \cite{chu2019famnet}                          & 52.0      & 14138     & 253616    & 3072      \\  
    \midrule[1pt]
    CPSN(Ours)                      & 52.1      & 16253     & 243672    & 3012      \\
    \bottomrule[2pt]
\end{tabular}
 \end{lrbox}
 
 \begin{table}[H]
 \caption{\small Comparison results among the proposed system and other multiple object tracking methods under MOTChallenge dataset. } 
 \vskip -0.1in
 \centering   
      \scalebox{0.85}{\usebox{\tabc}}
\label{tab:track_compare_motchallenge}
 \vskip -0.1in
 \end{table}  
 %%%%%%%%%%%%%%%%%%%%%%%%%%%%%%%%%%%%%%%%%%%%%%%%%%%%%%

\begin{figure}[t]
\centering
\includegraphics[width=\linewidth]{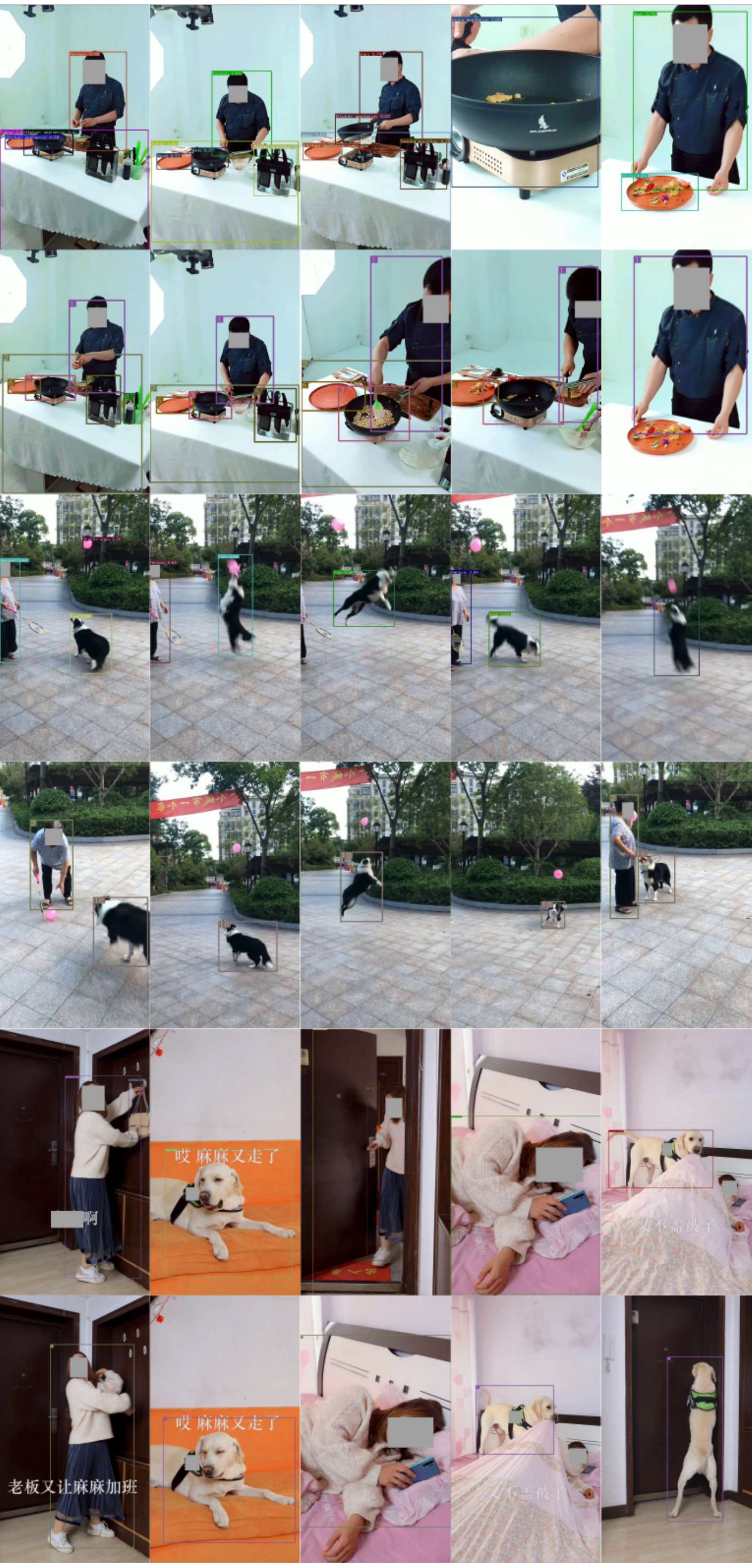}
\caption{\small Visualization of the detection results (odd rows) and re-identification results (even rows).}
\label{fig:result_traj}
\end{figure}

%%%%%%%%%%%%%%%%%%%%% 
\begin{lrbox}{\tabc}
\begin{tabular}{c|c|c|c|c|c}
    \toprule[2pt]
    \diagbox{Methods}{Metrics}      & MOTA      & FP        & FN        & IDS   & FPS           \\
    \midrule[1pt]
    Tracktor \cite{bergmann2019tracking}                     & 36.98     & 22.06     & 65.17     & 4.21  & 1.5           \\
    Tracktor++v2 \cite{bergmann2019tracking}                    & 41.56     & 20.01     & 64.16     & 4.01  & 1.5           \\
    GSM\_Tracktor \cite{liugsm}                   & 42.96     & 19.56     & \textbf{61.07}     & 3.90  & 8.7           \\
    MPNTrack \cite{braso2020learning}                        & \textbf{43.10}	    & \textbf{18.76}     & 61.52     & \textbf{3.54}  & 1.8           \\
    \midrule[1pt]
    CPSN(Ours)                      & 41.85     & 20.16     & 63.06     & 3.94  & \textbf{12.94}\\
    \bottomrule[2pt]
\end{tabular}
 \end{lrbox}
%%%%%%%%%%%%%%%%%%%%%%%%%%

 %\begin{table}[htbp]
 \begin{table}[t]
 \caption{\small Comparison results among the proposed system and other object tracking methods under the proposed SVD-IOR dataset. } 
 \vskip -0.1in
 \centering   
      \scalebox{0.85}{\usebox{\tabc}}
\label{tab:track_compare_svd}
 \vskip -0.1in
 \end{table}  
%%%%%%%%%%%%%%%%%%%%%%%%%%%%%%%%%%%%%%%%%%%%%%%%%%%%%%

On the contrary, Table \ref{tab:track_compare_svd} shows that our system achieves higher MOTA than Tracktor and Tracktor++v2 in short video dataset, which are the state-of-the-art video object tracking methods available in 2019. It indicates that our system has promising performance in video object tracking. Moreover, the proposed CPSN keep a higher efficiency than most of those methods.

\begin{figure}[t]
\centering
\includegraphics[width=\linewidth]{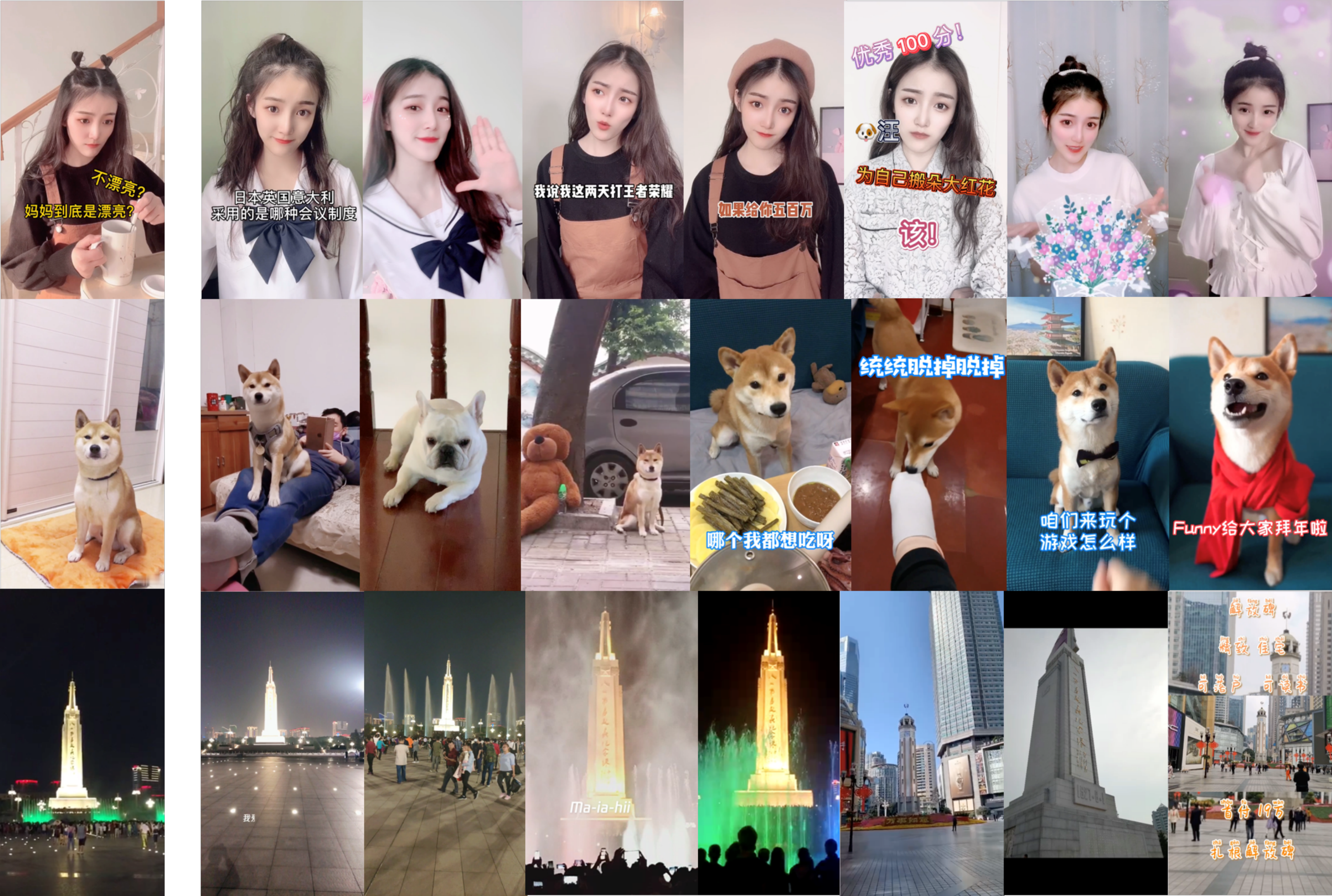}
\caption{\small Visualization of the re-identification results among different short videos. The first column denotes the query short videos, and the following columns denote query results in order of retrieve confidence.}
\label{fig:result_reid}
\end{figure}

\subsection{Evaluation of Re-Identification Module}
For identical object trajectory extraction among different video shot in single short videos, the detection and re-identification results under the proposed SVD-IOR dataset are visualized in Figure \ref{fig:result_traj}. As illustrated in the odd rows, objects with rare categories and complicated motion can be still well detected. The corresponding re-identification results in the even rows shows that the identical object can keep same track-ID among different short video frames and shots, due to the robust detection results and the proposed system.

%%%%%%%%%%%%%%%%%%%%% 
 \newsavebox{\tabd}
 \begin{lrbox}{\tabd}
\begin{tabular}{c|c|c|c|c|c|c}
    \toprule[2pt]
    
    \multirow{2}{*}{Methods}    &   \multirow{2}{*}{Rank-1 (\%)}     & \multirow{2}{*}{Rank-5 (\%)}   & \multicolumn{4}{c}{Rank-1 Accuracy of specific category (\%)}     \\
    \cline{4-7}
                                &                                    &                                & person    & cat    & dog    & landmark                              \\
    \midrule[1pt]
    baseline              & 58.5                               & 68.8                           & 75.0      & 36.5   & 52.5   & 70.0                               \\
    Ours      & \textbf{78.8}                               & \textbf{87.5}                           & \textbf{90.0}      & \textbf{75.0}   & \textbf{70.0}   & \textbf{80.0}                                 \\
    \bottomrule[2pt]
\end{tabular}
 \end{lrbox}
%%%%%%%%%%%%%%%%%%%%%%%%%%

 %\begin{table}[htbp]
 \begin{table}[t]
 \caption{\small Evaluation results of the proposed system on SOT-ReID dataset. } 
 \vskip -0.1in
 \centering   
      \scalebox{0.8}{\usebox{\tabd}}
\label{tab:reid}
 \vskip -0.1in
 \end{table}  
%%%%%%%%%%%%%%%%%%%%%%%%%%%%%%%%%%%%%%%%%%%%%%%%%%%%%%

For identical object retrieval among different videos, Table \ref{tab:reid} indicates that the proposed system achieves promising performance evaluated on SVD-ReID dataset. Compared with the baseline model, which uses YOLOv3-SPP in detection module and disables the query set update mechanism in re-identification module, the proposed system performs better in all $4$ categories. Due to the introduction of face detection and recognition, our system obtains high rank-1 accuracy in person re-identification among short videos. Moreover, landmark buildings are also well retrieved since each of them generally share a relatively consistent appearance in different videos. Cats and dogs in same breed but different ID are difficult to distinguish, as a result of which, these categories achieve a lower rank-1 accuracy. Re-identification results among different short videos are shown in Figure \ref{fig:result_reid}.

\section{Conclusion}
% Please check all the pages of your PDF file. Is the page size A4? Are there any type 3, Identity-H, or CID fonts? Are all the fonts embedded? Are there any areas where equations or figures run into the margins? Did you include all your figures? Did you follow mixed case capitalization rules for your title? Did you include a copyright notice? Do any of the pages scroll slowly (because the graphics draw slowly on the page)? Are URLs underlined and in color? You will need to fix these common errors before submitting your file. 
This paper makes two contributions to solving generic object re-identification problem for short videos. First, we propose a system composed of a detection module, a tracking module and a re-identification module, which formulate the challenging problem into three main sub-problems. In order to satisfy the high efficiency requested in practical application and get over the complicated variety of appearance of objects in short videos, we propose Temporal Information Fusion Network (TIFN) and Cross-layer Pointwise Siamese Network (CPSN) in detection module and tracking module respectively. Moreover, we propose two novel real-world short video datasets collected from short video platform for evaluating object trajectory extraction and generic object re-identification among different short videos. Quantitative experiments demonstrate the high effectiveness and efficiency of our system.

% \begin{acks}
% \end{acks}

\input{KDD_paper.bbl}

\bibliographystyle{ACM-Reference-Format}
% \bibliography{KDD_paper}

\end{document}